\documentclass{article}

\usepackage{xspace}
\usepackage{graphicx}
\usepackage{caption}
\usepackage{amsmath}
\usepackage{xcolor}
\usepackage{booktabs}

\usepackage{enumitem}

\definecolor{cvprblue}{rgb}{0.21,0.49,0.74}
\usepackage[pagebackref,breaklinks,colorlinks=true,linkcolor=cvprblue,citecolor=cvprblue,bookmarks=true]{hyperref}
\usepackage[capitalise,nameinlink]{cleveref}

\def\model{\textsc{SoccerBot}\xspace}

\makeatletter
\renewcommand\paragraph{\@startsection{paragraph}{4}{\z@}%
  {1.5ex \@plus 0.5ex \@minus 0.2ex}%
  {-0.8em}%
  {\normalfont\normalsize\bfseries}}
\makeatother

\usepackage[final]{corl_2026} 
\usepackage{amsmath}
\usepackage{amssymb}
\usepackage{booktabs}
\usepackage{graphicx}
\usepackage{xcolor}
\usepackage{url}

\title{Playing Soccer Like Humans Do: Learning Versatile and Agile Humanoid Soccer via Latent Motion Priors}

\author{
\textbf{Haoyu Shao$^{1,2}$, Yunshen Wang$^{2,3}$,
Jingze Zhang$^{2,4}$, Shaohang Zhu$^{1,2}$} \\
\textbf{Yong-Lu Li$^{3}$, Tianzhu Zhang$^{1}$, Baoxiong Jia$^{2,\dagger}$,
Siyuan Huang$^{2,\dagger}$} \\[0.5em]
$^{1}$ University of Science and Technology of China \\
$^{2}$ State Key Laboratory of General Artificial Intelligence, BIGAI \\
$^{3}$ Shanghai Jiao Tong University \quad
$^{4}$ Peking University \\
$^\dagger$ Corresponding authors.
}

\begin{document}
\maketitle

\begin{center}
    \centering
    \vspace{-25pt}
    \captionsetup{type=figure}
        \includegraphics[width=1\linewidth]{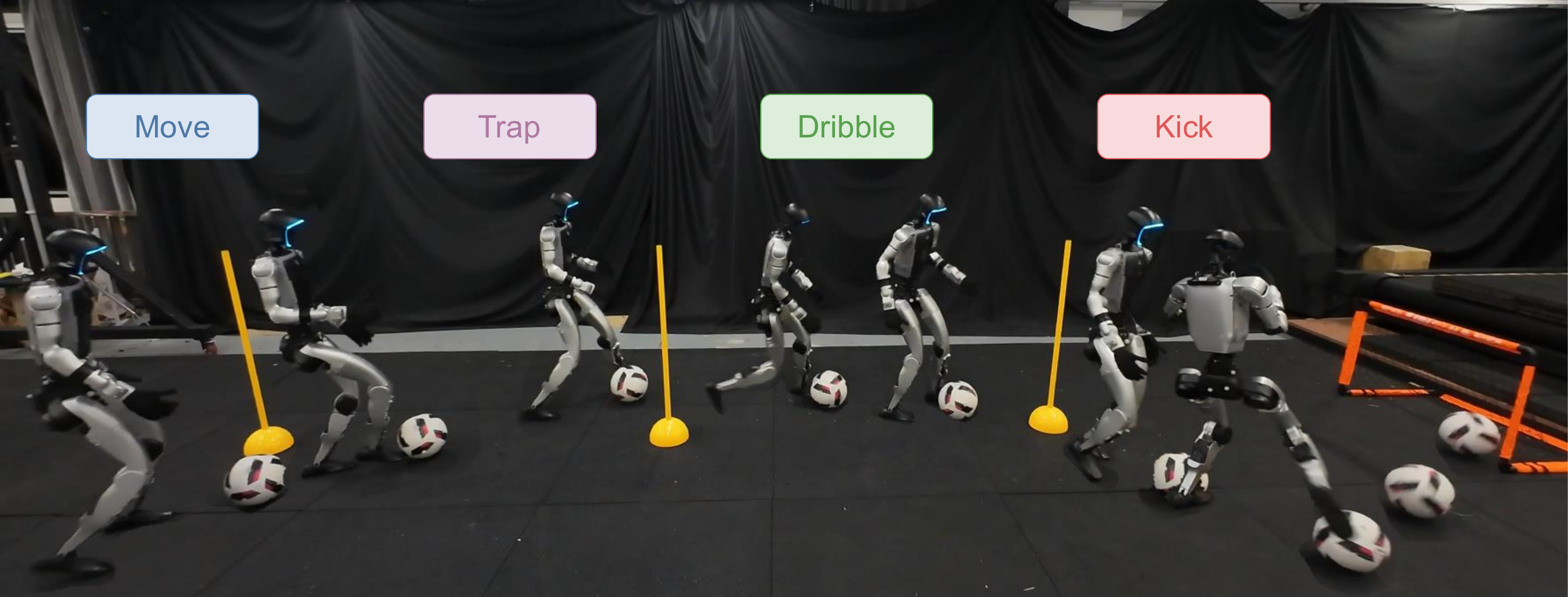}
        \captionof{figure}{
        \textbf{Real-world soccer skill execution on the Unitree G1 humanoid robot.} \model learns a single unified policy that covers the full soccer behavioral spectrum, including moving, trapping, dribbling, and kicking, with agile and human-like whole-body coordination. 
        }
    \label{fig:teaser}
\end{center}
\begin{abstract}
Humanoid soccer requires a robot to seamlessly combine agile locomotion, ball manipulation, and impact-rich contact dynamics across a continuous, dynamic behavioral spectrum. Yet existing approaches either cover only a limited set of skills, or forego motion priors entirely and produce unnatural behaviors, leaving the combination of a unified skill set and human-like motion quality unaddressed. We present \model, a hierarchical reinforcement learning framework that resolves both limitations through a structured latent motion prior distilled from large-scale human soccer motion capture data. By constraining policy learning within this dynamically plausible motion manifold, a single skill-conditioned policy naturally learns to perform moving, dribbling, trapping, and kicking in a unified formulation, while inherently producing human-like motion and fluid transitions during commanded skill switching. Augmented by robustness-oriented simulation design, \model transfers zero-shot to the Unitree G1 humanoid robot, demonstrating agile, natural, and robust execution of the full soccer skill repertoire under real-world conditions.

\end{abstract}

\keywords{Humanoid Soccer, Latent Motion Priors} 


\section{Introduction}
\label{sec:intro}
Soccer stands apart from sports such as table tennis~\cite{su2025hitter,ren2026smash} and tennis~\cite{zhang2026learningathletichumanoidtennis} as a uniquely demanding benchmark for humanoid robots: rather than isolating upper-body dexterity, it demands the inseparable coupling of agile whole-body locomotion and non-prehensile ball manipulation across a rich, continuous behavioral spectrum. Humans master this spectrum with remarkable fluency: sprinting, trapping, dribbling, and kicking compose seamlessly into a continuous, expressive behavioral repertoire. This fluency emerges from decades of motor learning that endows human motion with natural coordination, dynamic stability, and graceful skill transitions. Replicating such fluency on a humanoid robot is far from trivial: the robot must not only master a diverse set of highly dynamic skills, but also execute smooth, natural transitions across the full behavioral spectrum. Humanoid soccer thus emerges as a compelling platform for studying unified whole-body control, \textit{simultaneously demanding agile locomotion, non-prehensile object manipulation, impact-rich contact dynamics, and human-like motion quality within a single, coherent control problem.}

Despite recent progress in reinforcement learning for humanoid locomotion~\cite{radosavovic2024humanoid,radosavovic2024learning,radosavovic2024real} and motor skill acquisition~\cite{peng2021amp,peng2022ase,liao2025beyondmimic,ji2024exbody2,li2026omnitrack,yang2025omniretarget,wang2026omnixtreme,haarnoja2024learning}, existing approaches to humanoid soccer remain fundamentally limited along two orthogonal axes. On one axis, existing methods lack a unified skill repertoire: most prior works address only a narrow subset of soccer behaviors~\cite{kong2026learning,wang2026learning,wang2026dribble}, such as directed kicking, dribbling, or locomotion-only control, and even multi-skill systems rely on decoupled, independently trained controllers that cannot naturally bridge the dynamics between different actions~\cite{kim2025physicsfc}. Multi-skill soccer built on human motion priors has so far been demonstrated only for simulated characters~\cite{liu2022motor,kim2025physicsfc}. The result is a fragmented behavioral space where inter-skill transitions are inherently discontinuous and unnatural, as evidenced by the jarring switches between running and trapping, or dribbling and kicking. On the other axis, many existing methods rely solely on task-reward-driven reinforcement learning without any motion prior~\cite{haarnoja2024learning}, leaving motion naturalness entirely unaddressed. Without structural regularization grounded in human motion, such policies produce erratic, mechanically unnatural behaviors that are both visually implausible and brittle under the contact-rich dynamics of real soccer interaction. Together, these two axes define a clear gap in the literature: the combination of human-like coordination and a complete, unified soccer skill set has never been simultaneously achieved in a single humanoid soccer system.

To close this gap, we propose \model, a hierarchical learning framework that addresses both limitations through a unified mechanism: a structured latent motion prior distilled from large-scale human soccer motion capture data. Building on recent advances in physics-based motion imitation~\cite{luo2024universal}, we first distill a motion tracking policy into a compact latent action space that encodes a rich, dynamically plausible manifold of human-like whole-body motion. A unified skill-conditioned policy then operates within this latent space via residual control, learning to perform dribbling, trapping, kicking, and moving under a shared control formulation. By grounding all skills within the same motion prior, \model inherently produces natural, human-like motion across the full behavioral spectrum, and produces fluid, natural inter-skill transitions during commanded skill switching. To further support reliable real-world deployment, we incorporate robustness-oriented simulation design including ball dynamics modeling, domain randomization, and observation perturbation, enabling zero-shot sim-to-real transfer to the Unitree G1 humanoid robot.

We evaluate \textsc{SoccerBot} in both simulation and real-world settings, demonstrating that a single unified policy can execute the full spectrum of soccer skills with agility and naturalness under real-world conditions. The main contributions of this work are as follows:

\begin{itemize}[leftmargin=*,noitemsep,nolistsep,topsep=0pt,partopsep=0pt]
\item We present \model, to our knowledge the first real-world humanoid soccer system that jointly achieves a complete skill repertoire and human-like motion quality, learning moving, dribbling, trapping, and kicking within a single unified policy.
\item We propose a hierarchical latent motion prior framework with transition state initialization (TSI), grounding unified soccer skill learning within a structured motion manifold and enabling agile, human-like behaviors, fluid inter-skill transitions, and robust cross-skill composition without explicit state machines or separate controllers.
\item We conduct extensive simulation benchmarks and real-world experiments on the Unitree G1, demonstrating strong task performance across all four skills, structured latent skill-space organization, and reliable zero-shot sim-to-real transfer with minimal performance degradation.
\end{itemize}
Videos are available at \url{https://soccerbot-humanoid.github.io}.

\section{Related Work}
\label{sec:related_work}
\paragraph{Humanoid control and athletic skills.}
Reinforcement learning has driven rapid progress in humanoid robot control, enabling robust locomotion~\cite{radosavovic2024real,radosavovic2024humanoid,radosavovic2024learning} and whole-body motion tracking~\cite{liao2025beyondmimic,ji2024exbody2,chen2025gmt,zhang2025track,wang2026omnixtreme,li2026omnitrack} from large-scale human motion capture data~\cite{mahmood2019amass}. Beyond locomotion and motion tracking, RL has enabled humanoid robots to master a growing range of athletic skills, including tennis~\cite{zhang2026learningathletichumanoidtennis}, table tennis~\cite{su2025hitter,ren2026smash}, and parkour~\cite{wu2026perceptive,yang2025omniretarget}. Among these, humanoid soccer stands out as uniquely demanding: it requires the tight coupling of agile locomotion and non-prehensile ball manipulation through the feet, sustained across a continuous and dynamic behavioral spectrum. Early learning-based soccer systems demonstrated agile ball interaction on simulated avatars~\cite{liu2022motor} and small real bipedal robots~\cite{haarnoja2024learning}, while more recent work has explored soccer motion generation from human demonstrations~\cite{vahl2025soccerdiffusion, yang2025smgdiff}. On real humanoid robots, Dribble Master~\cite{wang2026dribble} learns agile dribbling through a two-stage curriculum with virtual-camera perception, and other works study perception-driven kicking and reactive soccer skills~\cite{kong2026learning,wang2026learning}. Most closely related to our work, PhysicsFC~\cite{kim2025physicsfc} learns dribbling, trapping, moving, and kicking with smooth transitions, but relies on skill-specific policies with an explicit finite-state machine and remains confined to simulation. In contrast, our work learns a single unified policy across the full soccer skill repertoire and deploys it on a real humanoid robot.
\paragraph{Latent motion priors for humanoid control.}
A prominent approach to regularizing humanoid reinforcement learning uses adversarial motion priors. AMP~\cite{peng2021amp} learns a discriminator from unstructured motion clips to provide a style reward that guides task policies toward natural behaviors, and ASE~\cite{peng2022ase} extends this by learning reusable adversarial skill embeddings for downstream task policies. While effective, adversarial approaches require careful balancing between task reward and motion realism, and can struggle to discover precise contact-rich strategies such as ball trapping.
An alternative line of work constructs structured latent action spaces by distilling motion tracking controllers into compact motor representations. PULSE~\cite{luo2024universal} distills a universal humanoid motion tracker into a VAE-style latent action space with a learned state-conditioned prior, enabling downstream policies to generate coherent human-like motions without adversarial rewards. Inspired by this framework, recent work has extended latent motion priors to athletic humanoid skills, demonstrating their effectiveness for contact-rich tasks such as tennis~\cite{zhang2026learningathletichumanoidtennis}. We adapt the latent motion prior framework to unified humanoid soccer, training a single residual latent policy conditioned on skill commands across moving, dribbling, trapping, and kicking, which extends beyond single-skill athletic tasks to a complete soccer skill repertoire deployed on a real robot.

\begin{figure}
    \centering
    \includegraphics[width=1.0\linewidth]{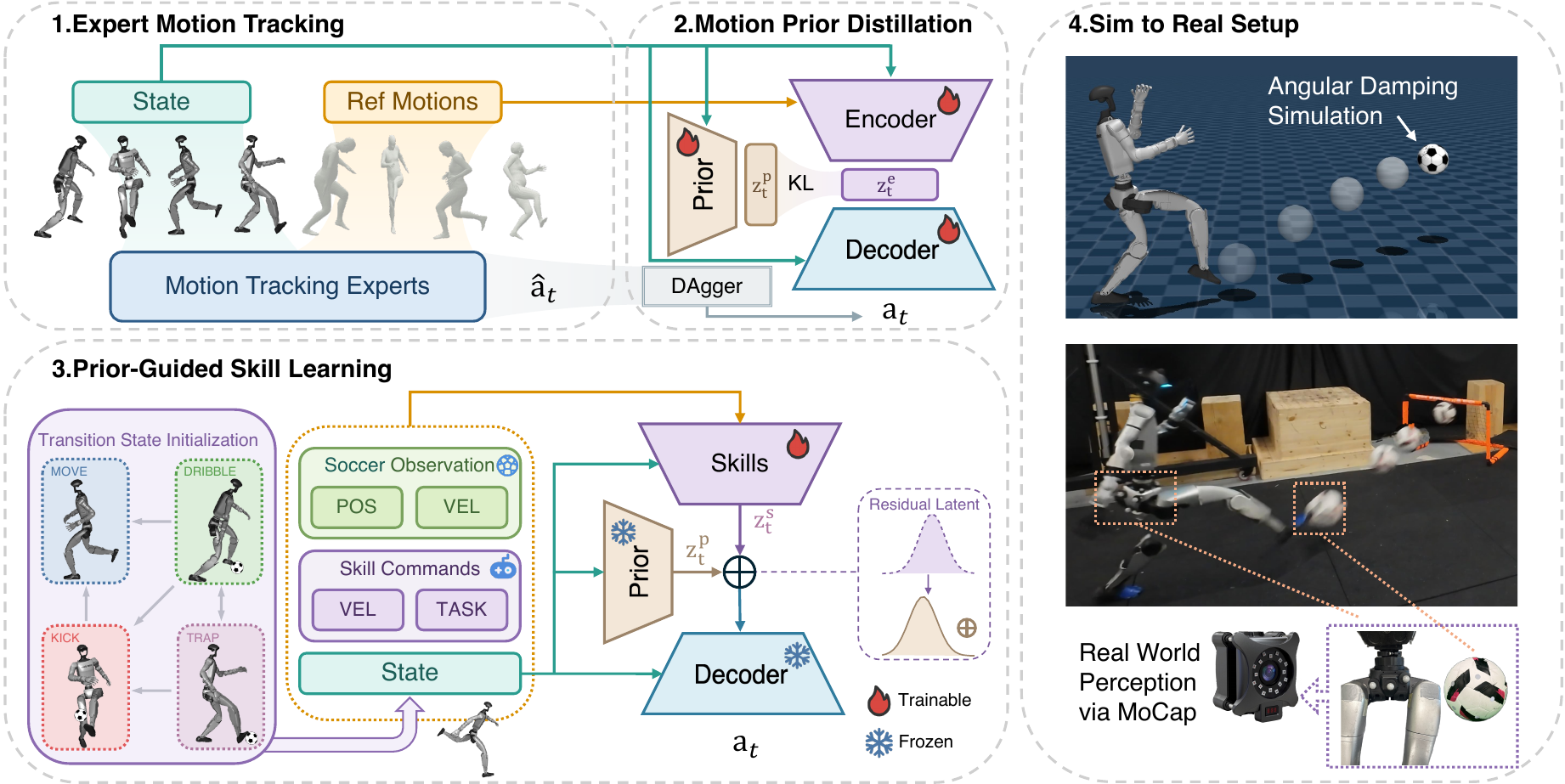}
    \caption{\textbf{Overview of \model.} The framework consists of three stages: (1) a motion tracking expert is trained on retargeted human soccer motion data; (2) the tracking expert is distilled into a structured latent motion prior via a variational encoder-decoder framework; (3) a unified skill-conditioned policy learns to perform moving, dribbling, trapping, and kicking via residual latent control over the frozen prior, with transition state initialization enabling robust cross-skill composition. Real-world deployment uses an external motion capture system for ball state estimation.}
    \label{fig:method}
    \vspace{-5pt}
\end{figure}

\section{Method}
\label{sec:method}

We present \textsc{SoccerBot}, a staged framework for learning a unified whole-body policy for humanoid soccer skills. The key idea is to separate low-level motor competence from high-level soccer task optimization. As shown in~\cref{fig:method}, our pipeline has three stages: Stage~I trains a motion tracking teacher from curated soccer demonstrations; Stage~II distills this teacher into a compact latent motion prior; and Stage~III trains a skill-conditioned soccer policy within this frozen latent action space. The final policy receives a skill command and task observation, predicts a latent residual, and decodes it into joint position targets through the pretrained motion decoder.

\subsection{Motion Tracking Policy Learning}
\label{sec:tracking}

The first stage constructs a physically feasible motion corpus and trains a whole-body tracking policy $\pi_{\mathrm{trk}}$. We source human soccer demonstrations from Soccer-X~\cite{yang2025smgdiff} and retarget them to the humanoid robot using GMR~\cite{joao2025gmr}. Since raw retargeted motions may contain kinematic artifacts such as foot skating, discontinuous contacts, and dynamically infeasible poses, we apply a filtering pipeline: a pretrained tracker rolls out each sequence in simulation, and segments that cannot be tracked under hardware torque limits are discarded. This yields approximately $2.5$ hours of curated reference motions $\mathcal{D}_{\mathrm{motion}}$.

We train $\pi_{\mathrm{trk}}(a_t\mid s_t,\hat{s}_{\mathrm{ref}})$ with reinforcement learning to imitate the filtered trajectories. The state is $s_t=(s_t^{\mathrm{prop}},s_t^{\mathrm{priv}})$, where $s_t^{\mathrm{prop}}$ contains proprioceptive observations such as joint positions and velocities, projected gravity, base angular velocity, and previous action, while $s_t^{\mathrm{priv}}$ denotes privileged information available only in simulation, such as the base linear velocity. The reference $\hat{s}_{\mathrm{ref}}\in\mathcal{D}_{\mathrm{motion}}$ specifies target full-body poses and velocities. The policy outputs joint position targets tracked by a low-level PD controller, and the reward combines pose and velocity imitation with balance and control regularization. After convergence, $\pi_{\mathrm{trk}}$ is executed on $\mathcal{D}_{\mathrm{motion}}$ and provides the teacher actions for distillation.

\subsection{Latent Motion Prior Construction via Distillation}
\label{sec:distillation}

The second stage distills the tracking teacher into a structured latent motion prior. The student consists of a posterior encoder $q_{\phi}(z_t\mid s_t^{\mathrm{prop}},s_t^{\mathrm{ref}})$, a state-conditioned prior $p_{\psi}(z_t\mid s_t^{\mathrm{prop}})$, and a decoder $D_{\theta}(z_t,s_t^{\mathrm{prop}})$. The posterior observes both proprioception and reference motion, while the prior observes only proprioception. This asymmetry lets the encoder infer the latent that best explains the teacher motion, while forcing the prior to model the motions that are feasible from the current body state alone, which is exactly the information available to the downstream soccer policy.

We model both distributions as diagonal Gaussians,
\[
q_{\phi}=\mathcal{N}(\mu_{\phi}(s_t^{\mathrm{prop}},s_t^{\mathrm{ref}}),\mathrm{diag}(\sigma_{\phi}^2)),
\qquad
p_{\psi}=\mathcal{N}(\mu_{\psi}(s_t^{\mathrm{prop}}),\mathrm{diag}(\sigma_{\psi}^2)).
\]
Given the teacher action $a_t^\star=\pi_{\mathrm{trk}}(s_t,\hat{s}_{\mathrm{ref}})$, the student samples
$z_t=\mu_{\phi,t}+\sigma_{\phi,t}\odot\epsilon_t$, $\epsilon_t\sim\mathcal{N}(0,I)$, and reconstructs
$\hat{a}_t=D_{\theta}(z_t,s_t^{\mathrm{prop}})$. Training uses an online DAgger-style distillation buffer~\cite{ross2011reduction} containing student-visited states and the corresponding teacher actions.

The distillation loss is
\begin{equation}
\mathcal{L}_{\mathrm{distill}}
=
\mathbb{E}_t\|\hat{a}_t-a_t^\star\|_2^2
+
\beta_{\mathrm{KL}}\mathbb{E}_t D_{\mathrm{KL}}(q_{\phi}\|p_{\psi})
+
\beta_{\mathrm{sm}}\mathbb{E}_t\|\mu_{\phi,t}-\mu_{\phi,t-1}\|_2^2 .
\label{eq:distillation_loss}
\end{equation}
The first term trains the decoder to reproduce the teacher's joint targets. The second term regularizes the full information posterior toward the learned prior, so that the information carried by the reference motion is transferred into a distribution that can be queried without it. The third term penalizes abrupt changes of the posterior mean and yields temporally smooth latent trajectories, which improves the continuity of the decoded joint targets. After distillation, the posterior is discarded, while $p_{\psi}$ and $D_{\theta}$ are frozen for downstream soccer learning.

\subsection{Skill-Conditioned Soccer Policy Learning}
\label{sec:soccerskills}

The third stage trains a unified skill-conditioned policy $\pi_\eta$ for the soccer skills. A single policy handles four primitives: \texttt{dribbling}, \texttt{kicking}, \texttt{moving}, and \texttt{trapping}. Its observation contains the velocity command, ball position and velocity in the robot frame, skill one-hot vector, and proprioceptive state. For \texttt{moving}, ball observations are masked, so the same interface supports both ball-centric and robot-centric behaviors without skill-specific observation spaces.

\subsubsection{Latent Space Residual Actor}

To preserve the stability and naturalness of the motion prior during dynamic ball interaction, the soccer policy does not act directly in joint space. Instead, it learns a residual in the pretrained latent motion space while the prior and decoder stay frozen. Given proprioception $s_t^{\mathrm{prop}}$, task observation $s_t^{\mathrm{task}}$, and skill command $e_t$, the frozen prior produces $z_t^p=\mu_{\psi}(s_t^{\mathrm{prop}})$. The policy samples a residual
$z_t^s\sim\mathcal{N}(\mu_\eta(s_t^{\mathrm{prop}},s_t^{\mathrm{task}},e_t),\mathrm{diag}(\sigma_\eta^2))$ and decodes
\begin{equation}
    z_t=z_t^p+\alpha z_t^s,
    \qquad
    a_t=D_\theta(z_t,s_t^{\mathrm{prop}}).
    \label{eq:latent_residual_policy}
\end{equation}
Here $\alpha$ scales the residual and $\sigma_\eta$ is a learnable state-independent standard deviation. PPO optimizes this latent residual distribution rather than the $29$-dimensional joint action: likelihood ratios, entropy, and KL terms are computed only in latent space. This restricts stochastic exploration to the neighborhood of the motion prior while still allowing the policy to adapt to dynamic ball interactions.

We regularize the residual mean by
\begin{equation}
\mathcal{L}_{\mathrm{bound}}
=
\mathbb{E}_t
\sum_i
\left[
\max\left(|\mu_{\eta,i}(s_t^{\mathrm{prop}},s_t^{\mathrm{task}},e_t)|-1,0\right)
\right]^2 .
\label{eq:bound_loss}
\end{equation}
This keeps the task policy close to the motion prior while preserving the expressivity needed for ball interaction.

\subsubsection{Multi-Skill Training and Transition Curricula}
\label{sec:skill_conditioned_training}

Each episode samples an active skill and interprets the command accordingly. For \texttt{moving}, the command specifies the desired robot velocity. For the ball-centric skills, it specifies the desired ball motion: \texttt{dribbling} samples low-speed target ball velocities, \texttt{kicking} samples high-speed target velocities together with strikeable initial ball states, and \texttt{trapping} launches incoming ground passes that the robot must intercept and damp. Rewards are routed by skill: only the terms of the active skill are enabled at each timestep, while a shared set of regularizers encourages consistent whole-body behavior.

\paragraph{Transition State Initialization.}
We use \emph{transition state initialization} (TSI) to train robust skill switching. For each skill, we keep a circular buffer of terminal robot and ball states from episodes that ended without failure. At reset, an episode of the target skill can be initialized from the terminal state of a preceding skill. This exposes the policy to the postures, momentum, and ball configurations resulting from the preceding skill, rather than relying solely on canonical initial states.

\subsection{Robustness for Sim-to-Real Transfer}
\label{subsec:sim2real}

We train with lightweight randomization and observation corruption that target the two main sources of sim-to-real mismatch in soccer: contact-rich robot--ball dynamics and imperfect state estimation.

\paragraph{Physics and dynamics randomization.}
We randomize robot--ground and ball--ground friction and restitution and the ball mass, and we apply angular damping to the ball with a coefficient $c_a$ sampled uniformly from $[0.3,0.6]$ for each environment at initialization. This encourages repeated interaction with the ball to maintain the commanded speed during dribbling. We further randomize the torso center of mass, default joint positions, and actuator latency. External velocity pushes during rollouts improve recovery from contact-induced disturbances.

\paragraph{Observation robustness.}
Policy observations are corrupted with bounded noise on ball position, ball velocity, base angular velocity, projected gravity, joint positions, and joint velocities, which makes the controller less dependent on exact simulator states.

\section{Experimental Results}
\label{sec:result}
The multi-skill policy is trained within the NVIDIA Isaac Lab simulation environment tailored to the Unitree G1 humanoid robot. For real-world hardware deployment, exteroceptive states are captured via an external optical motion capture system. The tracked ball states are transformed into the local robot base coordinate frame before being fed into the control policy.

\subsection{Main Simulation Results on Soccer Skills}
\paragraph{Experiment Setting} We evaluate the proposed framework on four humanoid soccer skills: kicking, dribbling, trapping, and moving. These skills require not only task completion, but also stable whole-body dynamic balancing and human-like motion quality. Each method is evaluated on $2{,}500$ episodes per skill with randomized simulation settings.

\paragraph{Baselines and Ablation Study}
We compare our method with the following baselines:
1) \textbf{PPO}: vanilla PPO trained with our task reward without human motions. 
2) \textbf{AMP}: PPO with an adversarial motion prior reward computed from human motions. 
3) \textbf{ASE}: an adversarial skill embedding baseline that pre-trains a latent-conditioned low-level controller on the human motions and then trains a high-level task policy that selects skill latents. 
4) \textbf{LATENT}: a task policy trained in a learned latent action space with Latent Action Barrier~\cite{zhang2026learningathletichumanoidtennis}. 
We compare our method with the following ablations: 
1) \textbf{Ours w/o prior}: removes the learned state-conditioned prior and predicts task latents directly. 
2) \textbf{Ours w/o TSI}: trains without TSI.

\paragraph{Metrics}
We use a compact set of task and motion metrics. \textbf{SR} is the percentage of trials that satisfy the skill-specific success condition without a fall or task failure. \textbf{VelErr} measures the speed tracking error of the task object: the ball for kicking and dribbling, and the robot base for moving. \textbf{DirErr} is the heading error between the realized motion and the commanded direction in the horizontal plane. For dribbling, \textbf{ResT} is the response time after a command switch. For trapping, \textbf{StopT} is the time from first ball contact to a stable stop, and \textbf{DistErr} is the ball displacement during stopping. \textbf{Train ref.} and \textbf{Post-train ref.} measure how closely a policy moves like a human: each is the squared maximum mean discrepancy, MMD$^2$, between motion features of the policy's rollouts and those of a human reference set. Train ref.\ uses the training motions $\mathcal{D}_{\mathrm{motion}}$, and Post-train ref.\ consists of motion capture clips that were not used to train any of the evaluated methods.

\paragraph{Experiment Results}
Table~\ref{tab:main_skills} summarizes performance across kicking, dribbling, trapping, and moving. Ours achieves the highest kicking ($97.04\%$) and dribbling ($89.88\%$) SR of all methods and completes every moving command. PPO kicks and traps reliably but never dribbles successfully, AMP fails most trapping and dribbling episodes, and ASE learns neither kicking nor dribbling.

\paragraph{Motion Alignment with Human References}
Every method that uses human motion data is closer to the training motions than PPO, which shares our task rewards but uses none. AMP, which directly rewards matching the motion distribution, is closest to both reference sets, and ASE, whose low-level controller imitates the reference motions, is second on the post-training clips. However, AMP traps only $23.08\%$ of passes and dribbles in $46.36\%$ of episodes, and ASE learns neither kicking nor dribbling. Among the methods that complete all four skills, Ours is closest to both reference sets.
\paragraph{Ablations}
The components serve distinct roles: the frozen decoder, distilled from human soccer motions, carries the motion representation in every latent variant; the state-conditioned prior supplies a feasible latent anchor around which the residual explores; and TSI broadens the distribution of switching states. Because the w/o prior variant keeps the same decoder, it isolates the anchor. Both ablations trap more reliably than Ours. Ours tends to block fast passes with an extended leg, and the ball then rebounds beyond the displacement gate. Removing the prior lowers kicking and dribbling SR, and removing TSI lowers dribbling SR. The prior shows instead in how the policy moves: with the same decoder, removing it moves the policy further from both human reference sets, raising the mean MMD$^2$ to $0.234$ and $0.582$. The benefit of TSI appears mainly when skills start from non-canonical states (\cref{sec:transition,sec:kick_robustness}).

\begin{table*}[t]
\centering
\caption{Main simulation results on four humanoid soccer skills.}
\label{tab:main_skills}
\resizebox{\linewidth}{!}{
\begin{tabular}{l ccc cccc ccc ccc cc}
\toprule
Method &
\multicolumn{3}{c}{Kicking} &
\multicolumn{4}{c}{Dribbling} &
\multicolumn{3}{c}{Trapping} &
\multicolumn{3}{c}{Moving} &
\multicolumn{2}{c}{Motion alignment} \\
\cmidrule(lr){2-4}
\cmidrule(lr){5-8}
\cmidrule(lr){9-11}
\cmidrule(lr){12-14}
\cmidrule(lr){15-16}

& SR $\uparrow$ & VE $\downarrow$ & DE  $\downarrow$
& SR $\uparrow$ & VE $\downarrow$ & DE  $\downarrow$ & ResT $\downarrow$
& SR $\uparrow$ & StopT $\downarrow$ & DistErr $\downarrow$
& SR $\uparrow$ & VE $\downarrow$ & DE $\downarrow$
& Train ref. $\downarrow$ & Post-train ref. $\downarrow$\\
\midrule

PPO   & 93.96 & 8.68 & 4.35 & 0.00 & 99.81 & 163.83 & 5.00 & 94.64 & 0.87 & 0.31 & 99.96 & 6.07 & 1.77 & 0.328 & 0.575 \\
AMP~\cite{peng2021amp}   & 87.48 & 10.32 & 7.26 & 46.36 & 37.05 & 47.88 & 4.09 & 23.08 & 5.49 & 3.07 & 99.96 & 9.43 & 4.21 & 0.170 & 0.308 \\
ASE~\cite{peng2022ase}   & 0.00 & 99.52 & 169.79 & 0.00 & 99.67 & 163.64 & 5.00 & 49.24 & 3.46 & 1.58 & 58.60 & 28.59 & 26.62 & 0.292 & 0.409 \\
LATENT~\cite{zhang2026learningathletichumanoidtennis} & 91.00 & 8.50 & 6.80 & 85.24 & 12.69 & 7.21 & 1.92 & 85.92 & 1.53 & 0.64 & 100.00 & 9.67 & 2.71 & 0.231 & 0.603 \\
\midrule
\textbf{Ours} & 97.04 & 7.26 & 4.76 & 89.88 & 8.79 & 4.24 & 1.36 & 83.08 & 1.43
  & 0.49 & 100.00 & 4.95 & 5.38 & 0.226 & 0.554 \\
-- w/o prior & 94.40 & 8.47 & 4.79 & 89.48 & 7.67 & 3.06 & 1.38 & 94.00 &
  0.97 & 0.32 & 100.00 & 4.67 & 2.87 & 0.234 & 0.582 \\
-- w/o TSI & 95.80 & 7.81 & 5.23 & 87.76 & 13.68 & 4.24 & 1.48 & 98.12 & 0.72 & 0.13 & 100.00 & 4.79 & 2.10 & 0.252 & 0.587 \\
\bottomrule
\end{tabular}}
\end{table*}

\subsection{Skill Transition and Composability}
\label{sec:transition}

\paragraph{Experiment Setting}
We evaluate whether the four learned skills can naturally switch between each other. For each source skill, we roll out and store terminal robot and ball states. Each target skill is then evaluated from these states. For ball-centric targets, only source states with the ball within $0.8$\,m of the robot are eligible. We compare the full model with the variant trained without TSI.

\begin{figure*}[t]
\centering

\begin{minipage}[t]{0.55\linewidth}
\centering
\captionof{table}{Terminal state skill transition success rate (\%). Each cell reports \textbf{Ours / w/o TSI}.}
\label{tab:transition_matrix}
\resizebox{\linewidth}{!}{
\begin{tabular}{lcccc}
\toprule
From / To & Dribble & Kick & Move & Trap  \\
\midrule
Dribble & 94.0 / 90.0 & 87.5 / 78.0 & 93.0 / 91.5 & 53.0 / 71.0 \\
Kick    & -- / -- & -- / -- & 99.5 / 97.5 & 86.5 / 96.5 \\
Move    & 78.5 / 71.5 & -- / -- & 92.5 / 91.5 & 58.5 / 68.5 \\
Trap    & 87.5 / 85.0 & 96.5 / 92.0 & 99.0 / 100.0 & 79.5 / 95.5 \\
\bottomrule
\end{tabular}}
\end{minipage}
\hfill
\begin{minipage}[t]{0.43\linewidth}
\centering
\captionof{table}{Kicking success rate (\%) across controlled ball and command conditions.}
\label{tab:kick_condition_eval}
\resizebox{\linewidth}{!}{
\begin{tabular}{lcccc}
\toprule
Method & Near & Mid & Incoming & WideYaw \\
\midrule
Ours w/o prior & 97.4 & 76.5 & 92.4 & 88.2  \\
Ours w/o TSI & 95.3 & 34.3 & 60.4 & 73.7 \\
Ours(sim) & 98.6 & 87.0 & 72.9 & 83.9 \\
\midrule
Ours(real) & 95.0 & 85.0 & 70.0 & 80.0 \\
\bottomrule
\end{tabular}}
\end{minipage}

\vspace{0.4cm}

\includegraphics[width=0.95\linewidth]{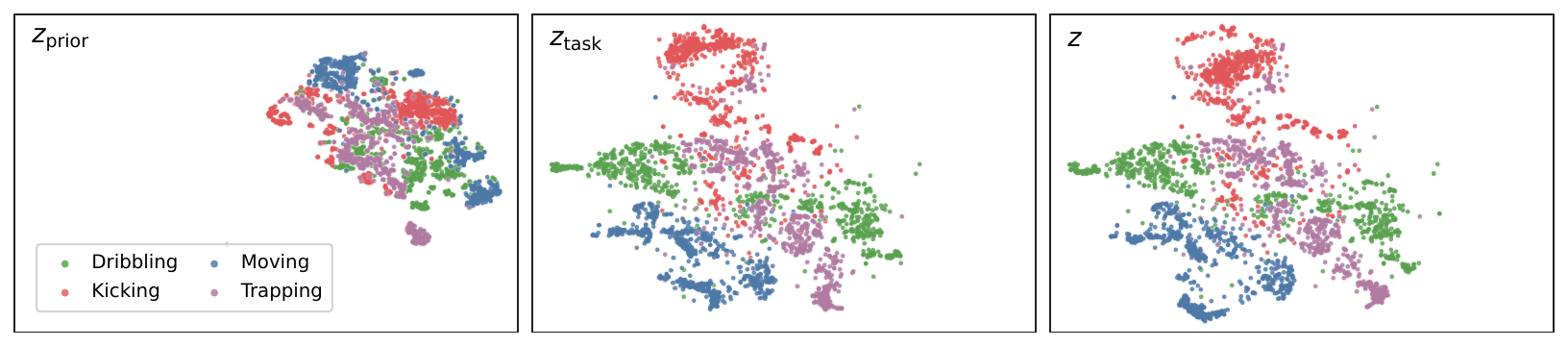}
\caption{
Latent skill-space construction under a joint t-SNE projection.
Samples from $z_{\mathrm{prior}}$, $z_{\mathrm{task}}$, and the final latent $z$ are embedded together and shown in separate panels.
Colors indicate skill labels.
The prior latent captures shared state-conditioned motor modes, while the task residual and final latent reveal increasingly skill-dependent structure.
}
\label{fig:latent_analysis}
\end{figure*}

\paragraph{Metrics}
Each entry in Table~\ref{tab:transition_matrix} is the SR of the target skill when initialized from terminal states of the source skill. Off-diagonal entries evaluate cross-skill switching, while diagonal entries evaluate command resampling within the same skill.

\paragraph{Experiment Results}
Table~\ref{tab:transition_matrix} shows that the learned skills can be composed from terminal states of other skills. For non-trapping targets, the full model is higher than the variant without TSI in $8$ of $9$ cells, most on Dribble$\rightarrow$Kick ($87.5\%$ vs.\ $78.0\%$), Move$\rightarrow$Dribble ($78.5\%$ vs.\ $71.5\%$), and Trap$\rightarrow$Kick ($96.5\%$ vs.\ $92.0\%$), whereas from canonical starts the two differ by only $1$--$2$ points on kicking and dribbling. For trapping targets, the variant without TSI is higher in all four cells, mirroring the full model's weaker trapping of fast passes from canonical starts rather than a transition-specific effect.

\subsection{Kicking Robustness Across Ball and Command Conditions}
\label{sec:kick_robustness}

\paragraph{Experiment Setting and Metrics}
We evaluate the kicking policy across four controlled settings: \textit{Near} and \textit{Mid} static balls with different initial distances, \textit{Incoming} balls moving toward the robot, and \textit{WideYaw} commands with large target yaw angles. 
The command speed is fixed to $8.0\,\mathrm{m/s}$, and each setting is tested over $1000$ deterministic simulation episodes. 
We report the success rate (\textbf{SR}), where success requires valid ball contact, satisfying the kicking velocity and direction gates, and maintaining balance.

\paragraph{Experiment Results}
Table~\ref{tab:kick_condition_eval} summarizes the kicking success rates under different ball and command conditions. 
In simulation, our full method achieves strong performance on near and mid-range static balls, with success rates of $98.6\%$ and $87.0\%$, respectively. 
The policy also remains robust under more challenging incoming ball and wide yaw settings, achieving $72.9\%$ and $83.9\%$ success rates. 
Compared with the variant without TSI, the full method improves all four conditions by $3.3$--$52.7$ points, most strongly on mid-range balls, suggesting that the more diverse reset distribution helps the policy adapt to kicking situations away from the nominal setup. 

The real-world results further demonstrate the sim-to-real transferability of the learned policy in controlling kicking direction, with success evaluated solely based on directional accuracy.
Despite the hardware and perception uncertainties in real deployment, the real-robot success rates remain close to the simulation results, with only a small average performance drop of about $3.1$ percentage points. 
This suggests that the learned kicking strategy is not only effective in simulation but also transfers well to the real robot across different ball and command conditions.

\subsection{Latent Skill-Space Construction}

\paragraph{Experiment Setting}
We analyze how the latent action is constructed from the motion prior and the task residual. During evaluation rollouts, we record the prior latent $z_{\mathrm{prior}}$, the task residual latent $z_{\mathrm{task}}$, and the final latent $z$ for kicking, dribbling, moving, and trapping. Samples from all three latent types are concatenated and embedded with a single joint t-SNE projection, then shown as three separate panels in Fig.~\ref{fig:latent_analysis}. Since the three panels share the same embedding space, we compare local neighborhoods and skill clusters, rather than interpreting the t-SNE axes directly.

\paragraph{Experiment Results}
Fig.~\ref{fig:latent_analysis} shows a transition from a shared motion prior to a skill-aware latent space. In $z_{\mathrm{prior}}$, the four skills largely overlap, suggesting that the prior captures state-conditioned feasible motor modes rather than explicit skill identity. The residual latent $z_{\mathrm{task}}$ introduces clearer skill-dependent separation. The final latent $z$ preserves the shared support of the prior while forming structured regions for dribbling, kicking, moving, and trapping. Moving and dribbling remain partially close because both rely on locomotion dynamics, while kicking and trapping occupy more specialized regions associated with striking and stopping the ball. This visualization indicates that the policy builds a structured skill space by adapting a shared motion prior with skill-conditioned residuals, rather than learning four isolated controllers.

\subsection{Real-World Evaluation and Sim-to-Real Analysis}
\label{sec:real}

\paragraph{Experiment Setting}
We deploy the policy zero-shot on a Unitree G1 humanoid robot. Ball and robot states are tracked by an external optical motion capture system and transformed into the robot base frame. We evaluate each of the four skills with $20$ trials: moving and dribbling vary the commanded speed and heading, trapping varies the incoming ball velocity, and kicking varies the target ball speed and heading. Table~\ref{tab:real_four_skills} compares it with its sibling simulation policy.

\paragraph{Experiment Results}
Table~\ref{tab:real_four_skills} shows the sim-to-real gap. Moving transfers most closely in speed: its speed error rises only from $19.14\%$ in simulation to $21.53\%$ on hardware. Dribbling and trapping degrade more across their continuous metrics. For dribbling, the speed error grows from $8.55\%$ to $30.17\%$ and the response time from $1.21$\,s to $2.37$\,s. For trapping, the stopping time grows from $0.78$\,s to $1.84$\,s, and the ball travels $1.07$\,m after contact instead of $0.31$\,m. Both skills depend on sustained robot--ball contact and rolling dynamics. Kicking retains substantially better directional than speed control: DirErr increases from $4.09^\circ$ to $10.41^\circ$, whereas VelErr increases from $8.12\%$ to $47.38\%$. 

\paragraph{Sim-to-Real Analysis}
To diagnose the remaining gap, we conducted two ball dynamics identification experiments on the deployment surface. Twenty passive drop trials yield a ball--ground restitution of $e=0.654$, which falls within the randomized range of $[0.45,0.85]$. Free rolling trials measure an effective rolling deceleration of $0.100$\,m/s$^2$. Reproducing these trials in simulation over the randomized angular damping range yields mean decelerations of $0.097$ to $0.167$\,m/s$^2$, so both measurements fall within the corresponding domain randomization ranges, supporting coverage of the observed real-world ball dynamics. The higher outgoing ball speed on hardware points to stronger effective foot--ball impulse transfer: the simulated ball is rigid, whereas the real inflated ball deforms during impact, altering the contact duration, force profile, and effective restitution.

\begin{table}[t]
\centering
\caption{Simulation and real-robot results on four humanoid soccer skills.}
\label{tab:real_four_skills}
\resizebox{\linewidth}{!}{
\begin{tabular}{lccccccccc}
\toprule
Method &
\multicolumn{2}{c}{Kicking} &
\multicolumn{3}{c}{Dribbling} &
\multicolumn{2}{c}{Trapping} &
\multicolumn{2}{c}{Moving} \\
\cmidrule(lr){2-3}
\cmidrule(lr){4-6}
\cmidrule(lr){7-8}
\cmidrule(lr){9-10}
& VelErr (\%) & DirErr ($^\circ$)
& VelErr (\%) & DirErr ($^\circ$) & ResT (s)
& StopT (s) & DistErr (m)
& VelErr (\%) & DirErr ($^\circ$) \\
\midrule
\textbf{Ours (sim)}
& 8.12 & 4.09
& 8.55 & 3.77 & 1.21
& 0.78 & 0.31
& 19.14 & 5.85 \\
\textbf{Ours (real)}
& 47.38 & 10.41
& 30.17 & 16.49 & 2.37
& 1.84 & 1.07
& 21.53 & 13.74 \\
\bottomrule
\end{tabular}}
\end{table}


\section{Conclusion}
\label{sec:conclusion}

We presented \textsc{SoccerBot}, a hierarchical learning framework for unified humanoid soccer control. By decoupling whole-body motor competence from task-level optimization, our approach distills human motion data into a reusable latent motion prior, upon which a single skill-conditioned policy learns moving, dribbling, kicking, and trapping within a shared latent action space. Simulation and real-world experiments on the Unitree G1 demonstrate robust skill execution, natural whole-body motion, and stable sim-to-real transfer. Beyond individual skills, the unified command interface enables higher-level planners to compose learned primitives into longer-horizon behaviors, suggesting a practical path toward autonomous planning and potentially multi-robot coordination.

\section{Limitations and Future Work}
\label{sec:limitations}
Despite promising results, several limitations remain. First, the gap between simulation and real deployment varies across skills. On hardware, kicks overshoot the commanded speed, while dribbling and trapping exhibit the largest increases in continuous error metrics. These discrepancies may partly stem from the rigid ball contact model. Future work could explore compliant contact modeling and policy fine-tuning on the real robot. Second, the system relies on an external motion capture system for state estimation, restricting deployment to instrumented indoor environments. Integrating onboard perception would enable deployment in less structured settings. Third, the skill set is limited to four fundamental soccer behaviors; extending it with broader motion data and task curricula would make the policy more expressive. Fourth, our experiments focus on single-robot control, leaving multi-agent coordination and game-level decision making for future work. A natural direction is to use the learned skill-conditioned policy as a reusable low-level interface for autonomous planning and multi-agent humanoid soccer.

\clearpage
\acknowledgments{We thank the area chair and the anonymous reviewers for their constructive feedback, which helped improve this paper.}


\bibliography{example}

\clearpage
\appendix
\label{sec:appendix}

\setcounter{figure}{0}
\setcounter{table}{0}
\setcounter{equation}{0}
\renewcommand{\thefigure}{A.\arabic{figure}}
\renewcommand{\thetable}{A.\arabic{table}}
\renewcommand{\theequation}{A.\arabic{equation}}

\section{Problem Formulation}
\label{sec:appendix:formulation}
We formulate humanoid soccer skill learning as a skill-conditioned Markov decision process
$\mathcal{M}=(\mathcal{S},\mathcal{A},\mathcal{C},p,r,\gamma)$. At time $t$, the state
$s_t\in\mathcal{S}$ contains the robot and ball states, the action
$a_t\in\mathcal{A}$ is a vector of joint position targets, and $c_t\in\mathcal{C}$ denotes the active skill. The skill set is
\[
\mathcal{C}=\{\texttt{dribbling},\texttt{kicking},\texttt{moving},\texttt{trapping}\}.
\]
Given the observation $o_t$, the objective is
\begin{equation}
    \max_{\pi}
    \mathbb{E}_{c\sim p(c),\,\tau\sim\pi(\cdot\mid c)}
    \left[
        \sum_{t=0}^{T-1}\gamma^t r_c(s_t,a_t)
    \right],
    \qquad
    a_t=\pi(o_t,c),
    \label{eq:appendix:mdp_objective}
\end{equation}
where $r_c$ is a skill-conditioned reward. The final soccer policy does not act directly in the raw joint action space. Instead, it predicts a residual latent action, combines it with a frozen state-conditioned motion prior, and decodes the resulting latent into joint position targets.

\subsection{Observations and Commands}
\label{sec:appendix:observations}
The actor observes the command, masked ball state, skill one-hot code, proprioception, and previous action. For \texttt{moving}, ball position and velocity are masked to zero so that the shared policy interface does not depend on irrelevant ball state. The critic uses the same clean terms and additionally observes base linear velocity. Table~\ref{tab:appendix_observations} gives the complete observation layout, dimensions, and policy-side corruption ranges.

\begin{table*}[t]
\centering
\caption{Observation layout for the unified soccer skill policy and critic. The policy observation concatenates all listed policy terms and applies corruption where specified. The critic uses the corresponding uncorrupted terms, additionally includes base linear velocity, and disables corruption. For the Unitree G1, there are 29 actuated joints, yielding 106 policy dimensions and 109 critic dimensions.}
\label{tab:appendix_observations}
\resizebox{\linewidth}{!}{
\begin{tabular}{llcl}
\toprule
Group & Term & Dim. & Corruption \\
\midrule
Policy/Critic & Command & 3 & None \\
Policy/Critic & Ball position & 3 & $[-0.02,0.02]$ m \\
Policy/Critic & Ball velocity & 3 & $[-0.1,0.1]$ m/s \\
Policy/Critic & Skill one-hot & 4 & None \\
Policy/Critic & Base angular velocity & 3 & $[-0.05,0.05]$ rad/s \\
Critic only   & Base linear velocity  & 3 & None \\
Policy/Critic & Projected gravity & 3 & $[-0.02,0.02]$ \\
Policy/Critic & Relative joint position & 29 & $[-0.01,0.01]$ rad \\
Policy/Critic & Relative joint velocity & 29 & $[-0.5,0.5]$ rad/s \\
Policy/Critic & Last action & 29 & None \\
\bottomrule
\end{tabular}}
\end{table*}

\section{Reward Specification}
\label{sec:appendix:reward_specification}
Rewards are routed by skill so that only the objective associated with the active soccer skill is optimized at each timestep:
\begin{equation}
    r_t
    =
    r_t^{\mathrm{shared}}
    +
    \sum_{c\in\mathcal{C}}\mathbf{1}_{\{c_t=c\}}r_t^c .
    \label{eq:appendix:reward_routing}
\end{equation}
This routing avoids conflicting gradients between objectives, such as accelerating the ball for \texttt{kicking} and damping the ball for \texttt{trapping}. Shared terms regularize posture, action smoothness, joint limits, and termination behavior. Table~\ref{tab:appendix_rewards} lists the implemented task and regularization terms with their weights. Compact gates $g(\cdot)$ denote implementation-specific conditions, such as possession, contact phase, standing-command mode, and reward ramping. 

\begin{table*}[t]
\centering
\caption{Reward terms for the unified skill environment. $I_{\mathrm{drib}}$, $I_{\mathrm{kick}}$, $I_{\mathrm{move}}$, and $I_{\mathrm{trap}}$ denote active-skill indicators; $v_c$, $v_b$, and $v_r$ denote the command, ball, and root velocities, respectively; $p_b^r$ denotes the ball position in the robot frame; $d_f$ denotes the nearest foot--ball distance; and $g(\cdot)$ denotes implementation-specific gates, including contact phase, possession, standing-command mode, and reward ramping.}
\label{tab:appendix_rewards}
\resizebox{\linewidth}{!}{
\begin{tabular}{lp{10.4cm}r}
\toprule
Reward term & Expression & Weight \\
\midrule
\texttt{dribble\_ball\_vel} &
$I_{\mathrm{drib}}\exp\{-10[(\|v_c-v_b\|/(\|v_c\|+\epsilon))^2+0.1((\|v_c\|-\|v_b\|)/(\|v_c\|+\epsilon))^2]\}\,g_{\mathrm{poss}}g_{\mathrm{ramp}}$ & 8.0 \\
\texttt{dribble\_ball\_root\_pos} &
$I_{\mathrm{drib}}\exp(-\|p_b^r-[0.4,0]\|^2/0.2^2)\,g_{\mathrm{front}}$ & 2.0 \\
\texttt{dribble\_root\_vel} &
$I_{\mathrm{drib}}\exp\{-10[(\|v_r-v_c^\rightarrow\|/(\|v_c\|+\epsilon))^2+0.1((\|v_c\|-\|v_r\|)/(\|v_c\|+\epsilon))^2]\}$ & 2.0 \\
\texttt{dribble\_stand\_still} &
$I_{\mathrm{drib}}I_{\|v_c\|<0.05}I_{\|v_b\|<0.05}(1-\exp[-c_{\mathrm{still}}])$ & -6.0 \\
\midrule
\texttt{kick\_soccer\_vel} &
$I_{\mathrm{kick}}I_{\mathrm{contact}}I_{\|v_c\|>0.1}\,r_{\mathrm{speed}}r_{\mathrm{dir}}r_{\mathrm{upright}}$ & 10.0 \\
\texttt{kick\_no\_lateral\_vel} &
$I_{\mathrm{kick}}I_{\mathrm{contact}}I_{\|v_c\|>0.1}\,\mathrm{clip}(v_b^\parallel/\|v_c\|,0,1)\exp[-8(v_b^\perp/\|v_c\|)^2]$ & 2.0 \\
\texttt{kick\_approach} &
$I_{\mathrm{kick}}I_{\neg\mathrm{contact}}I_{\|v_c\|>0.1}(0.45r_{\mathrm{behind}}+0.35r_{\mathrm{foot}}+0.20r_{\mathrm{approach}})g_{\mathrm{align}}$ & 3.0 \\
\texttt{kick\_align} &
$I_{\mathrm{kick}}g_{\mathrm{pre/post}}g_{\mathrm{close}}\exp[2(\hat d_{\mathrm{ball}}\cdot\hat d_{\mathrm{root}}-1)]$ & 1.0 \\
\texttt{kick\_stand} &
$I_{\mathrm{kick}}g_{\mathrm{contact/zero}}\exp[-(\|v_r\|^2+0.5\|\omega_r\|^2+0.05\|\dot q\|^2+0.5\|q-q_0\|^2)]$ & 10.0 \\
\midrule
\texttt{move\_track\_vel} &
$I_{\mathrm{move}}\exp\{-4[(\|v_c-v_r\|/(\|v_c\|+\epsilon))^2+0.1((\|v_c\|-\|v_r\|)/(\|v_c\|+\epsilon))^2]\}$; for $\|v_c\|<0.05$: $I_{\mathrm{move}}\exp[-(\|v_r\|/0.03)^2]$ & 7.0 \\
\texttt{move\_track\_dir} &
$I_{\mathrm{move}}\exp(-2\theta^2)$, $\theta=\arccos(\hat v_c\cdot\hat d_{\mathrm{root}})$ & 2.0 \\
\texttt{move\_turn\_stability} &
$I_{\mathrm{move}}I_{\|v_c\|>0.05}\exp[-1.5\max(0,|\omega_z|-0.6-2\theta/\pi)^2-0.4(\omega_x^2+\omega_y^2)]$ & 1.5 \\
\midrule
\texttt{trap} &
$I_{\mathrm{trap}}\,[I_{\neg\mathrm{contact}}(0.1r_{\mathrm{foot-ball}}+0.3r_{\mathrm{approach}}+0.1r_{\mathrm{height}})+I_{\mathrm{contact}}(0.9r_{\mathrm{stop}}+0.1r_{\mathrm{step}}+1)]$ & 1.5 \\
\texttt{trap\_align} &
$I_{\mathrm{trap}}\exp[3(\hat d_{\mathrm{ball}}\cdot\hat d_{\mathrm{root}}-1)]$ & 1.0 \\
\texttt{trap\_post\_contact\_still} &
$I_{\mathrm{trap}}I_{\mathrm{contact}}g_{\mathrm{ramp}}\exp[-(\|v_r\|/0.12)^2]\exp[-(\|v_b\|/0.08)^2]$, $g_{\mathrm{ramp}}$ rising from $0.12$ to $0.5$\,s after contact & 0.25 \\
\midrule
\texttt{action\_rate\_l2} &
$\|a_t-a_{t-1}\|^2$ & -0.02 \\
\texttt{foot\_landing\_impact} &
$g_{\mathrm{skill}}\min\{6,\sum_{f}I_{\mathrm{land},f}[(\max(0,F_f-300)/600)^2+0.35\max(0,-\dot z_f-0.35)^2]\}$ & -0.01 \\
\texttt{joint\_limit} &
$p_{\mathrm{joint\ limit}}(q)$ & -20.0 \\
\texttt{joint\_pos\_upper\_body} &
$\|q_{\mathrm{upper}}-q_{0,\mathrm{upper}}\|^2$ & -0.1 \\
\texttt{termination\_penalty} &
$I_{\mathrm{terminated}}(1-I_{\mathrm{timeout}})$ & -200.0 \\
\texttt{torso\_upright\_reward} &
$\exp[-10\|z_{\mathrm{torso}}^{xy}\|^2]$ & 0.5 \\
\texttt{stand\_still\_reward} &
$I_{\|v_c\|<0.05}g_{\mathrm{ball}}\exp[-(\|v_r\|^2+0.5\|\omega_r\|^2+0.05\|\dot q\|^2+0.5\|q-q_0\|^2)]$ & 2.0 \\
\bottomrule
\end{tabular}}
\end{table*}

\section{Training, Transition State Initialization, and Robustness}
\label{sec:appendix:training_robustness}
We train the soccer actor with PPO in Isaac Lab/Isaac Sim using $8192$ parallel environments. The actor samples residual latents, and the frozen decoder maps them to joint position targets tracked by the robot low-level controller. The critic can use privileged simulation information during training, whereas the deployed actor only requires the policy observation terms in Table~\ref{tab:appendix_observations}. 

\subsection{Transition State Initialization}
\label{sec:appendix:tsi}
Transition state initialization (TSI) improves cross-skill recovery and skill switching. For each skill $c$, we maintain a circular buffer $\mathcal{B}_c$ of terminal states. A stored state contains the robot root state, joint positions and velocities, and the ball state. At reset, an episode starts either from the canonical task initialization or from a sampled terminal state:
\begin{equation}
    s_0\sim
    \begin{cases}
    p_{\mathrm{init}}(s_0\mid c), & \text{with probability }1-p_{\mathrm{TSI}}, \\
    \mathcal{B}_{c'}, & \text{with probability }p_{\mathrm{TSI}} .
    \end{cases}
    \label{eq:appendix:tsi_sampling}
\end{equation}
Here $c'$ is a predecessor skill of the sampled target skill $c$: dribbling can follow trapping; kicking can follow trapping and dribbling; moving can follow dribbling and kicking; and trapping can follow dribbling. The target command is then sampled and executed from the inherited robot and ball state. We use $p_{\mathrm{TSI}}=0.3$ and a buffer of $256$ states per skill. The buffer stores only terminal states of episodes that end without failure. This expands the reset distribution without injecting falls or physically invalid configurations.

\subsection{Domain Randomization and Observation Corruption}
\label{sec:appendix:domain_randomization}
The robustness design targets two deployment gaps: contact-rich robot--ball dynamics and noisy state estimation. Observation corruption is included directly in the \emph{Corruption} column of Table~\ref{tab:appendix_observations}. The implementation also supports a perception-stress setting that holds the previous ball observation and injects larger ball velocity noise for ball-interaction skills.

For dynamics mismatch, we randomize robot--ground and ball--ground friction and restitution, ball mass, ball angular damping, torso center of mass, default joint position offsets, actuator latency, external pushes, and initial reset states. Ball angular damping is applied by PhysX once per physics step ($\Delta t=5$\,ms):
\begin{equation}
    \boldsymbol{\omega}_{t+\Delta t}
    =
    (1-c_a\Delta t)\,\boldsymbol{\omega}_t,
    \label{eq:appendix:ball_angular_damping}
\end{equation}
where $c_a$ is the angular damping coefficient sampled once per environment. The full randomization ranges are summarized in Table~\ref{tab:appendix_domain_randomization}.

\begin{table*}[t]
\centering
\caption{Domain randomization and reset perturbations. Ranges are sampled uniformly unless otherwise stated.}
\label{tab:appendix_domain_randomization}
\resizebox{\linewidth}{!}{
\begin{tabular}{lll}
\toprule
Term & Unit & Range \\
\midrule
Robot static friction & -- & $[0.3,1.6]$ \\
Robot dynamic friction & -- & $[0.3,1.2]$ \\
Robot restitution & -- & $[0.0,0.15]$ \\
Soccer static friction & -- & $[0.35,0.75]$ \\
Soccer dynamic friction & -- & $[0.25,0.55]$ \\
Soccer restitution & -- & $[0.45,0.85]$ \\
Soccer angular damping & -- & $[0.3,0.6]$ \\
Soccer mass & kg & $[0.38,0.45]$ \\
Actuator latency & ms & $\{0,5,10\}$ \\
Push interval & s & $[1.0,3.0]$ \\
Default joint position offset & rad & $[-0.01,0.01]$ \\
Torso CoM offset, $x$ & m & $[-0.025,0.025]$ \\
Torso CoM offset, $y,z$ & m & $[-0.05,0.05]$ \\
Push linear velocity, $x,y$ & m/s & $[-0.5,0.5]$ \\
Push linear velocity, $z$ & m/s & $[-0.2,0.2]$ \\
Push angular velocity, roll/pitch & rad/s & $[-0.52,0.52]$ \\
Push angular velocity, yaw & rad/s & $[-0.78,0.78]$ \\
Initial root yaw & rad & $[-\pi,\pi]$ \\
Initial joint reset scale & -- & $[0.5,1.5]$ \\
\bottomrule
\end{tabular}}
\end{table*}

\end{document}